\documentclass[11pt,a4paper]{article}
\usepackage[hyperref]{emnlp2020}
\usepackage{times}
\usepackage{latexsym}
\usepackage{multirow}
\usepackage{booktabs}
\usepackage{tabularx}
\usepackage{graphics}
 \usepackage{graphicx}
 \usepackage{amssymb}
 \usepackage{subcaption}
\usepackage[ruled,vlined]{algorithm2e}

\usepackage{amsmath}
\newcommand{\hv}{{\boldsymbol h}}

\newcommand{\wv}{{\boldsymbol w}}

\newcommand{\sv}{{\boldsymbol s}}
\newcommand{\zv}{{\boldsymbol z}}
\newcommand{\Wmat}{{\bf W}}
\newcommand{\R}{\mathbb{R}}

\usepackage{url}

\aclfinalcopy 

\title{Multi-Fact Correction in Abstractive Text Summarization}

\author{Yue Dong$^1$\thanks{*Most of this work was done when the first author was
an intern at Microsoft.} \quad Shuohang Wang$^2$ \quad Zhe Gan$^2$ \quad Yu Cheng$^2$ \\ \textbf{Jackie Chi Kit Cheung}$^1$  \quad \textbf{Jingjing Liu}$^2$\\ \\
    $^1$Mila / McGill University, $^2$Microsoft Dynamics 365 AI Research \\
    {\small \{\tt yue.dong2@mail, jcheung@cs\}.mcgill.ca} \\
    \small \{\tt shuowa, zhe.gan, yu.cheng, jingjl \}@microsoft.com
    }

\date{}

\begin{document}
\maketitle

\begin{abstract}
  Pre-trained neural abstractive summarization systems have dominated extractive strategies on news summarization performance, at least in terms of ROUGE. However, system-generated abstractive summaries often face the pitfall of factual inconsistency: generating incorrect facts with respect to the source text. To address this challenge, we propose \textit{SpanFact}, a suite of two factual correction models that leverages knowledge learned from question answering models to make corrections in system-generated summaries via \textit{span selection}. Our models employ single or multi-masking strategies to either iteratively or auto-regressively replace entities in order to ensure semantic consistency w.r.t. the source text, while retaining the syntactic structure of summaries generated by abstractive summarization models. Experiments show that our models significantly boost the factual consistency of system-generated summaries without sacrificing summary quality in terms of both automatic metrics and human evaluation. 
\end{abstract}

\section{Introduction}

\begin{table}[t!]
\resizebox{\columnwidth}{!}{
\begin{tabular}{  m{5em} |  m{7cm}  } 
\hline
CNNDM Source& (CNN) \textcolor{blue}{About a quarter of a million} Australian homes and businesses have no power after \textcolor{blue}{a ``once in a decade" storm} battered Sydney and nearby areas. About 4,500 people have been isolated by flood waters as ``the roads are cut off and we won't be able to reach them for a few days,"...\\ 
\hline
 Bottom-up Summary& \textcolor{red}{a quarter of a million} australian homes and businesses have no power \textcolor{red}{after a decade}.\\
\hline
Corrected by SpanFact &\textcolor{orange}{about a quarter of a million} australian homes and businesses have no power \textcolor{orange}{after a ``once in a decade" storm.}\\
\hline \hline
Gigaword Source & all the \textcolor{blue}{12 victims} including \textcolor{blue}{8 killed and 4 injured} have been identified as senior high school students of the second senior high school of ruzhou city, central china's henan province, local police said friday. \\ 
\hline
 Pointer-Generator Summary& \textcolor{red}{12} killed, 4 injured in central china school shooting. \\ 
\hline
Corrected by SpanFact &\textcolor{orange}{8} killed, 4 injured in central china school shooting.\\
\hline \hline
XSum Source & st clare's \textcolor{blue}{catholic} primary school in birmingham has met with equality leaders at the city council to discuss a complaint from the pupil's family. the council is supporting the school to ensure its policies are appropriate...\\ 
\hline
 BertAbs Summary& a \textcolor{red}{muslim} school has been accused of breaching the equality act by refusing to wear headscarves. \\ 
\hline
Corrected by SpanFact &a \textcolor{orange}{catholic} school has been accused of breaching the equality act by refusing to wear headscarves.\\
\hline
\end{tabular}
}
\caption{Examples of factual error correction on different summarization datasets.  Factual
errors are marked in red. Corrections made by the proposed SpanFact models are marked in orange.}
\end{table}

Informative text summarization aims to shorten a long piece of text while preserving its main message. Existing systems can be divided into two main types: extractive and abstractive. Extractive strategies directly copy text snippets from the source to form summaries, while abstractive strategies generate summaries containing novel sentences not found in the source. Despite the fact that extractive strategies are simpler and less expensive, and can generate summaries that are more grammatically and semantically correct, abstractive strategies are becoming increasingly popular thanks to its flexibility, coherency and vocabulary diversity \citep{zhang2019pegasus}.

Recently, with the advent of Transformer-based models \citep{vaswani2017attention} pre-trained using self-supervised objectives on large text corpora \citep{devlin2018bert,radford2018gpt,lewis2019bart,raffel2019t5}, abstractive summarization models are surpassing extractive ones on automatic evaluation metrics such as ROUGE \citep{lin-2004-rouge}. However, several studies \cite{falke2019ranking,goodrich2019assessing,kryscinski2019evaluating,wang2020asking,durmus2020feqa,maynez2020faithfulness} observe that despite high ROUGE scores, system-generated abstractive summaries are often factually inconsistent with respect to the source text. Factual inconsistency is a well-known problem for conditional text generation, which requires models to generate readable text that is faithful to the input document. Consequently, sequence-to-sequence generation models need to learn to balance signals between the source for faithfulness and the learned language modeling prior for fluency \citep{kryscinski2019evaluating}. The dual objectives render abstractive summarization models highly prone to hallucinating content that is factually inconsistent with the source documents \citep{maynez2020faithfulness}. 

Prior work has pushed the frontier of guaranteeing factual consistency in abstractive summarization systems. Most focus on proposing evaluation metrics that are specific to factual consistency, as multiple human evaluations have shown that ROUGE or BERTScore \citep{zhang2020BERTScore} correlates poorly with faithfulness \citep{kryscinski2019evaluating,maynez2020faithfulness}.  These evaluation models range from using fact triples \cite{goodrich2019assessing},  textual entailment predictions \citep{ falke2019ranking}, adversarially pre-trained classifiers \citep{kryscinski2019evaluating}, to question answering (QA) systems \citep{wang2020asking,durmus2020feqa}. It is worth noting that QA-based evaluation metrics show surprisingly high correlations with human judgment on factuality \citep{wang2020asking}, indicating that QA models are robust in capturing facts that can benefit summarization tasks. 

On the other hand, some work focuses on model design to incorporate factual triples \citep{cao2018faithful,zhu2020boosting} or textual entailment \citep{li2018ensure,falke2019ranking} to boost factual consistency in generated summaries. Such models are efficient in boosting factual scores, but often at the expense of significantly lowering ROUGE scores of the generated summaries. This happens because the models struggle between generating pivotal content while retaining true facts, often with an eventual propensity to sacrificing informativeness for the sake of correctness of the summary. In addition, these models inherit the backbone of generative models that suffer from hallucination despite the regularization from complex knowledge graphs or text entailment signals. 

In this work, we propose SpanFact, a suite of two neural-based factual correctors that improve summary factual correctness without sacrificing informativeness. To ensure the retention of semantic meaning in the original documents while keeping the syntactic structures generated by advanced summarization models, we focus on factual edits on entities only, a major source of hallucinated errors in abstractive summarization systems in practice \citep{kryscinski2019evaluating,maynez2020faithfulness}. The proposed model is inspired by the observation that fact-checking QA model is a reliable medium in assessing whether an entity should be included in a summary as a fact \citep{wang2020asking,durmus2020feqa}. To our knowledge, we are the first to adapt QA knowledge to enhance abstractive summarization. Compared to sequential generation models that incorporate complex knowledge graph and NLI mechanisms to boost factuality, our approach is lightweight and can be readily applied to any system-generated summaries without retraining the model. Empirical results on multiple summarization datasets show that the proposed approach significantly improves summarization quality over multiple factuality measures without sacrificing ROUGE scores.

Our contributions are summarized as follows.
($i$) We propose SpanFact, a new factual correction framework that focuses on correcting erroneous facts in generated summaries, generalizable to any summarization system.
($ii$) We propose two methods to solve multi-fact correction problem with single or multi-span selection in an iterative or auto-regressive manner, respectively.
($iii$) Experimental results on multiple summarization benchmarks demonstrate that our approach can significantly improve multiple factuality measurements without a huge drop on ROUGE scores.

\section{Related Work}

\begin{figure*}[t!]
    \centering
    \includegraphics[width=\textwidth]{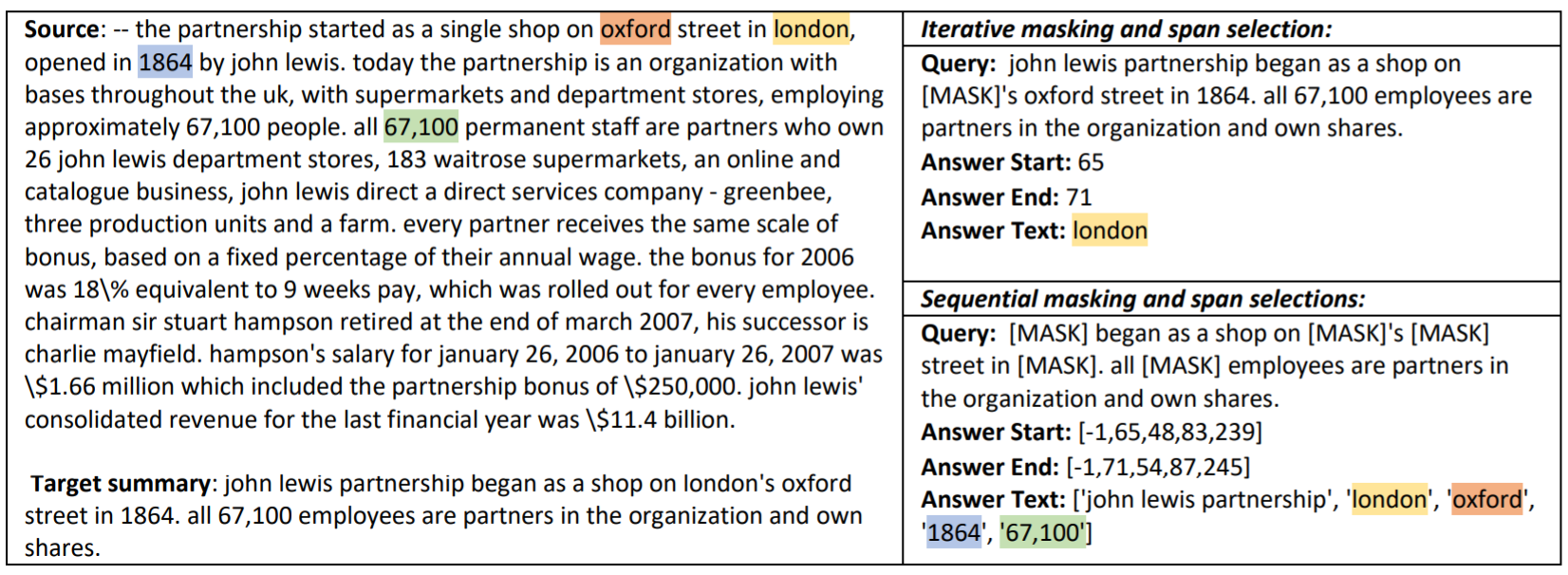}
    \caption{Training example created for the QA-span prediction model (upper right) and the auto-regressive fact correction model (bottom right).}
    \label{fig:span_selection_example}
\end{figure*}

The general neural-based encoder-decoder structure for abstractive summarization is first proposed by \citet{rush2015neural}. 
Later work improves this structure with better encoders, such as LSTMs \citep{chopra2016abstractive} and GRUs \citep{nallapati2016abstractive}, that are able to capture long-range dependencies, as well as with reinforcement learning methods that directly optimize summarization evaluation scores \citep{paulus2017deep}. One drawback of the earlier neural-based summarization models is the inability to produce out-of-vocabulary words, as the model can only generate whole words based on a fixed vocabulary. \citet{see2017get} proposes a pointer-generator framework that can copy words directly from the source through a pointer network \citep{vinyals2015pointer}, in addition to the traditional sequence-to-sequence generation model.

Abstractive summarization starts to shine with the advent of self-supervised algorithms, which allow deeper and more complicated neural networks such as Transformers \citep{vaswani2017attention} to learn diverse language priors from large-scale corpora. Models such as BERT \citep{devlin2018bert}, GPT \citep{radford2018gpt} and BART \citep{lewis2019bart} have achieved new state-of-the-art performances on abstractive summarization \citep{liu2019bertabs,lewis2019bart,zhang2019pegasus,shi-etal-2019-leafnats,fabbri-etal-2019-multi}. These models often finetune pre-trained Transformers with supervised summarization datasets that contain pairs of source and summary. 

However, encoder-decoder architectures widely used in abstractive summarization systems are inherently difficult to control and prone to hallucination \citep{vinyals2015neural,koehn2017six,lee2018hallucinations}, and often leads to factual inconsistency: the system-generated summary is fluent but unfaithful to the source \citep{cao2018faithful}. Studies have shown that  8\% to 30\% system-generated abstractive summaries have factual errors \citep{falke2019ranking,kryscinski2019evaluating} that cannot be discovered by ROUGE scores. Recent studies have proposed new methods to ensure factual consistency in summarization. \citet{cao2018faithful,zhu2020boosting} propose RNN-based and Transformer-based decoders that attend to both source and extracted knowledge triples, respectively. \citet{li2018ensure} propose an entailment-reward augmented maximum-likelihood training objective, and \citet{falke2019ranking} proposes to rerank beam results based on entailment scores to the source. 

\begin{figure*}[ht!]
    \centering
    \includegraphics[width=\textwidth]{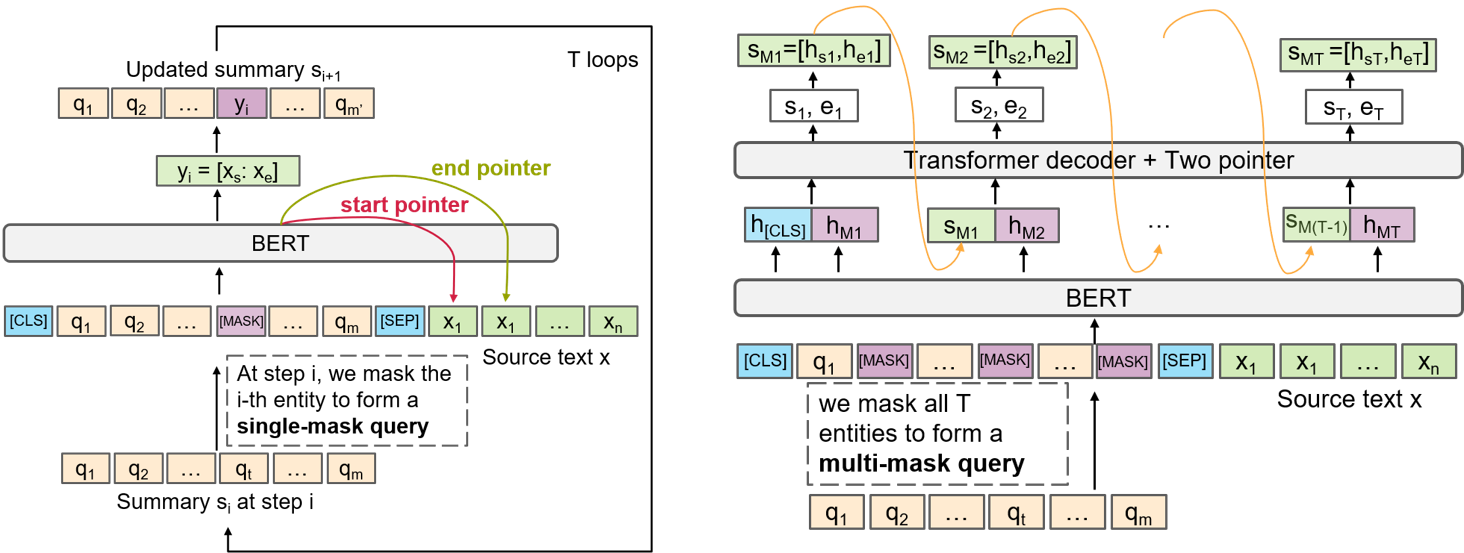}
    
    \caption{Model architecture (Left: QA-span fact correction model. Right: Auto-regressive fact correction model).}
    \label{fig:system}
\end{figure*}

Our fact correction models are inherently different from these models, as we focus on post-correcting summaries generated by any model. Our models are trained with the objective of predicting masked entities identified for fact correction (Figure \ref{fig:span_selection_example}), and learn to fill in the entity masks of any system-generated summaries with single or multi-span selection mechanism (Figure \ref{fig:system}). The most similar work to ours is proposed concurrently by \citet{cao-2020-fact}, where they fine-tune a BART \citep{lewis2019bart} model on distant supervision examples and use it as a post-editing model for factual error correction.

\section{Multi-Fact Correction Models}

In this section, we describe two models proposed for factual error correction: $(i)$ QA-span Fact Correction model, and $(ii)$ 
Auto-regressive Fact Correction model. As both methods rely on span selection with different masking and prediction strategies, we call them SpanFact collectively.

\subsection{Problem Formulation}
Let $(x, y)$ be a document-summary pair, where $x = (x_1, \ldots, x_M)$ is the source sequence with $M$ tokens, and $y = (y_1,\ldots, y_N )$ is the target sequence with $N$ tokens. An abstractive summarization
model aims to model the conditional likelihood $p(y|x)$, which can be factorized into a product $p(y|x) = \prod_{t=1}^{T} p(y_t|y_{1.\ldots,t-1}, x)$, where $y_{1.\ldots,t-1}$ denote the preceding tokens before position $t$. The conditional maximum-likelihood objective ideally requires summarization models to not only optimize for informativeness but also correctness. However, in reality this often fails as the models have a high propensity for leaning towards informativeness than correctness \citep{li2018ensure}. 

Suppose a summarization system generates a sequence of tokens $y' =(y'_1, \ldots, y'_N)$ to form a summary. Our factual correction models aim to edit an informative-yet-incorrect summary into $y'' =(y''_1, \ldots, y''_K)$ such that
\begin{align}
   f(x,y'') > f(x,y')\,, 
\end{align}
where $f$ is a metric measuring factual consistency between the source and system summary.

\subsection{Span Selection Dataset} \label{subsec:span_selection}

Our fact correction models are inspired by the \textit{span selection} task, which is often used in reading comprehension tasks such as question answering. Figure~\ref{fig:span_selection_example} shows examples of the \textit{span selection} datasets we created for training our QA-span and auto-regressive fact correction models, respectively. The \textit{query} is a reference summary masked with one or all entities,\footnote{In this work, we use SpaCy NER tagger \citep{honnibal2017spacy} to identify entities for data construction.} and the passage is the corresponding source document to be summarized. If an entity appears multiple times in the source document, we rank them based on the fuzzy string-matching scores (a variation of Levenshtein distance) between the query sentence and the source sentence containing the entity. Our models explicitly learn to predict the span of the masked entity rather than pointing to a specific token as in Pointer Network \citep{vinyals2015pointer}, because the original tokens and replaced tokens often have different lengths.

Our QA-span fact correction model iteratively mask and replace one entity at a time, while the auto-regressive model  masks  all the entities simultaneously, and replace them in an auto-regressive  fashion from left to right. Figure~\ref{fig:system} shows an overview of our models. Comparing the two models, the QA-span fact correction model works better when only a few errors exist in the draft summary, as the prediction of each mask is relatively independent of each other. 
On the other hand, the auto-regressive fact correction model starts with a skeleton summary that has all the entities masked, which is often more robust when summaries contain many factual errors.   

\subsection{QA-Span Fact Correction Model}
\label{subsec:iter_model}
In the iterative setting, our model aims to conduct entity correction by answering a query that contains only one mask at a time. Suppose a system summary has $T$ entities. At time step $i$, we mask the $i$-th entity and use this masked sequence as the query to our QA-span model. The prediction is placed into the masked slot in the query to generate an updated system summary to be used in the next step.  

Given the source text $x$ and a masked query $q = (y'_1, \ldots, \texttt{[MASK]},\ldots y'_m)$, our iterative correction model aims to predict the answer span via modeling $p(i=\textnormal{start})$ and $p(i=\textnormal{end})$. For span selection, we use the BertForQuestionAnswering\footnote{https://github.com/huggingface/transformers} model, which adds two separate non-linear layers on top of Transformers as pointers to the start and end token position for the answer. We initialize the fact-correction model from a pre-trained BERT model \citep{devlin2018bert}, and perform finetuning with the span selection datasets we created from the summarization datasets (Figure \ref{fig:span_selection_example}). 

The input to the BERT model is a concatenation of two segments: the masked query $q$ and the source $x$, separated by special delimiter markers as ($\texttt{[CLS]}, q, \texttt{[SEP]}, x$). Each token in the sequence is assigned with three embeddings: token embedding, position embedding, and segmentation embedding.\footnote{The segmentation embedding is used to distinguish the query (with two special tokens \texttt{[CLS]} and \texttt{[SEP]}) and the source in our models.} These embeddings are summed into a single vector and fed to the multi-layer Transformer model:
\begin{align}
    \tilde{\hv}^l &= \textnormal{LN}(\hv^{l-1}+\textnormal{MHAtt}(\hv^{l-1}))\,, \label{eq:transformer1} \\
    {\hv}^l &= \textnormal{LN}(\tilde{\hv}^{l}+\textnormal{FFN}(\tilde{\hv}^l))\,, \label{eq:transformer2}
\end{align}
where $\hv^0$ are the input vectors, and $l$ represents the depth of stacked layers. LN and MHAtt are layer normalization and multi-head attention operations \citep{vaswani2017attention}. The top layer provides the hidden states for the input tokens with rich contextual information. 
The start (s) and end (e) of the answer span are predicted as: 
\begin{align}
    a^{start}_i &=p(i=s) = \frac{\textnormal{exp}(q^s_i)}{\sum_{j =0}^{H-1} \textnormal{exp}(q^s_j)}\,, \label{eq:start1} \\
    a^{end}_i &=p(i=e) = \frac{\textnormal{exp}(q^e_i)}{\sum_{j = 0}^{H-1} \textnormal{exp}(q^e_j)}\,, \label{eq:end1} \\
  q^s_i &=\textnormal{ReLU}(\wv^\top_{s}\hv_i+b_{s})\,, \label{eq:start2} \\
    q^e_i &=\textnormal{ReLU}(\wv^\top_{e}\hv_i+b_{e})\,, \label{eq:end2}
\end{align}
where $H$ is the number of encoder's hidden states, $\wv_s,\wv_e \in \mathbb{R}^d$ and $b_s,b_e \in \mathbb{R}$ are trainable parameters. The final span is selected based on the argmax of Eqn. (\ref{eq:start1}) and (\ref{eq:end1}) with the constraint of  $p_{start}<p_{end}$ and  $p_{end}-p_{start}<k$.

\subsection{Auto-regressive Fact Correction Model}
One disadvantage of the QA-style span-prediction strategy is that if the sequence contains too many factual errors, masking out one entity at a time may lead to highly erroneous skeleton summary to start with. The model might be making predictions on top of wrong entities from later in the sequence. Masking one entity at a time is essentially a greedy local method that is prone to error accumulation. To alleviate this issue, we  
 propose a new sequential fact correction model to handle errors in a more global manner with beam search. Specifically, we mask out all the entities simultaneously, and use a novel \textit{auto-regressive} span-selection decoder to predict fillers for the multiple masks sequentially. By doing this, we assume dependency between the masks: the earlier predicted entities will be used as corrected context for better predictions in the later steps.

Given a source text $x = (x_1, \ldots, x_n)$ and a draft summary  $(y'_1,\ldots y'_m)$. Our model first masks out all the entities (with $T$ masks), and leaves a skeleton summary as the query  $q = (y'_1,\ldots, \texttt{[MASK]}_1,\ldots,\texttt{[MASK]}_T \ldots y'_m)$. Then, we concatenate the query $q$ with the source $x$  (similar to Section \ref{subsec:iter_model}) as inputs to the encoder. The inputs are fed into BERT to obtain contextual hidden representations. 

We then \textit{select} the encoder's hidden states for the $T$ masks $\hv_{y'_{mask_1}}, \ldots, \hv_{y'_{mask_T}}$ as partial input to an auto-regressive Transformer-based decoder. 
Unlike generation tasks that require an \texttt{[EOS]} token to indicate the end of decoding, our decoder runs $T$ steps to predict the answer spans for these $T$ masks. At step $t$, we first fuse the hidden representation $\hv_{\texttt{[MASK]}_t} \in \R^d$ of the $t$-th \texttt{[MASK]} token and previously predicted entity representation $\sv^{ent}_{t-1} \in \R^d$: 
\begin{equation}
    \label{eq:decoder_input}
    \zv_t = \Wmat [\hv_{\texttt{[MASK]}_t};\sv^{ent}_{t-1}]\,,
\end{equation}
where $\Wmat \in \mathbb{R}^{2d \times d}$, $\sv^{ent}_0 = \hv_{\texttt{[CLS]}}$ (the representation of \texttt{[CLS]} token), and $[;]$ denotes vector concatenation.

The input $\zv_t$ is then fed to the Transformer decoder (as in Eqn. (\ref{eq:transformer1}) and (\ref{eq:transformer2}))  to generate the decoder's hidden state $\hv'_t$ at time step $t$. Based on $\hv'_t$, we use a two-pointer network to predict the start and end positions of the answer entity in the source (encoder's hidden states). This is achieved with cross-attention of $\hv'_t$ w.r.t. the encoder's hidden states, similar to Eqn (\ref{eq:start1}) and (\ref{eq:end1}). This operation results in two distributions over the encoder's hidden states for the start and end span positions. The final prediction of the start and end positions for mask $t$ is obtained by taking the argmax\footnote{The argmax is used for selecting the start and end indexes for the answer span, and the softmax is used for computing the loss for back-propagation.} over the pointer position distributions:
\begin{align}
    p_{start} &= \arg\max(a^{start}_1,...,a^{start}_M)\,, \label{eq:argmax1} \\
    p_{end} &= \arg\max(a^{end}_1,...,a^{end}_M)\,, \label{eq:argmax2}
\end{align}
under the constraint that $p_{start}<p_{end}$ and  $p_{end}-p_{start}<k$.  

Based on the start and end positions for the predicted entity, we can obtain the predicted entity representation at time step $t$ as the mean over the in-span encoder's hidden states: 
\begin{equation}
    \label{eq:entity_forming}
    \sv^{ent}_t = \text{Mean-Pool}(\{\hv_{p_{start}}, \hv_{p_{end}}\})\,,
\end{equation}
which is used as the input for the next step of decoding. 
 It is worth noting that although the argmax operations in Eqn. (\ref{eq:argmax1}) and (\ref{eq:argmax2}) are non-differentiable, the model is trained based on the start and end positions of the ground-truth answer w.r.t. the start and end logits in Eqn. (\ref{eq:start1}) and (\ref{eq:end1}), which makes the gradient back-propagates to the encoder. Meanwhile, the encoder's hidden states used to compose  $\sv^{ent}_i$ in Eqn. (\ref{eq:entity_forming}) also carry the gradients. During inference, beam search is used to find the best sequence of predicted spans in the source to replace the masks. 
 
Compared to the conventional Pointer Network \citep{vinyals2015pointer,see2017get} that only points to one token at a time, our sequential span selection decoder has the flexibility to replace a mask by any number of entity tokens, which is often required in summary factual correction. 
\section{Experiment}
In this section, we present our results on using SpanFact for multiple summarization datasets.

\subsection{Experimental Setup}

Training data for our fact correction models are generated as described in Section \ref{subsec:span_selection} on CNN/DailyMail \cite{hermann2015teaching}, XSum \citep{narayan2018don} and Gigaword \citep{graff2003english,rush2015neural}. The statistics of these three dataset are provided in Table \ref{tab:datasets}.
During training, if an entity does not have a corresponding span in the source, we point the answer span to the \texttt{[CLS]} token. During inference, if the answer span predicted is the \texttt{[CLS]} token, we replace back the original masked entity. 

Our fact correction models are implemented via the Huggingface Transformers library \citep{Wolf2019HuggingFacesTS} in PyTorch \citep{paszke2017automatic}. We initialize all encoder models with the checkpoint of an uncased, large BERT model pre-trained on English data and SQuAD for all experiments. Both source and target texts were tokenized with BERT's sub-words tokenizer. The max sequence length is set to 512 for the encoder. We use a shallow Transformer decoder (L=2) for the auto-regressive span selection decoder, as the pre-trained BERT-large encoder is already robust for selecting right spans in the single-span selection task with only two pointers (Section \ref{subsec:iter_model}). The Transformer decoder has 1024 hidden units and the feed-forward intermediate size for all layers is 4,096.

All models were finetuned on our span prediction data for 2 epochs with batch size 12.  AdamW optimizer \citep{loshchilov2017decoupled} with $\epsilon=$1e-8 and an initial learning rate 3e-5 is used for training. Our learning rate schedule follows a linear decay scheduler with warmup=10,000. During inference, we use beam search with $b=5$ and $k=10$ (constraint for the distance between the start and end pointer). The best model checkpoints are
chosen based on performance on the validation set.  Experiments are conducted using 4 Quadro RTX 8000 GPUs with 48GB of memory.

\begin{table}[t!]
\resizebox{\columnwidth}{!}{%
\begin{tabular}{l|cccc}
\toprule
Datasets  & \# docs (train/val/test) & doc len. & summ. len. & \# mask \\ \midrule
CNN       & 90,266/1,220/1093        & 760.50          & 45.70  &     4.40      \\ 
DailyMail & 196,961/12,148/10,397    & 653.33          & 54.65  &    5.38       \\ 
XSum      & 204,045/11,332/11,334    & 431.07          & 23.26  &       2.28    \\ 
Gigaword  & 3,803,957/189,651/1,951  & 31.3            & 8.3    &     1.97      \\ \bottomrule
\end{tabular}
}
\caption{Comparison of summarization datasets on train/validation/test set splits, average document and summary length (numbers of words). We also report the average number of entity masks on the reference summary for each dataset.} \label{tab:datasets}
\end{table}

\subsection{Evaluation Metrics}
We use three automatic evaluation metrics to evaluate our models. The first is ROUGE \citep{lin-2004-rouge}, the standard summarization quality metric, which has high correlation with summary informativeness in the news domain \citep{kryscinski2019evaluating}. 

Since ROUGE has been criticized for its poor correlation with factual consistency  \citep{kryscinski2019evaluating,wang2020asking}, we use two additional automatic metrics that specifically focus on factual consistency: FactCC \citep{kryscinski2019evaluating} and QAGS  \citep{wang2020asking}. FactCC is a pre-trained binary classifier that evaluates the factuality of a system-generated summary by predicting whether it is consistent or inconsistent w.r.t. the source. This classifier was trained on adversarial examples obtained by heuristically injecting noise into reference summaries.

\begin{table}[t!]
\centering
\resizebox{\columnwidth}{!}{%
  \begin{tabular}{ l | c|c| c c c } 
    \toprule
     \multirow{2}{*}{Datasets} & QGQA & FactCC& \multicolumn{3}{c}{ROUGE} \\
    & & sent & 1 & 2 & L \\ \midrule
    Bottom-up & 70.58 & 73.66  &\textbf{41.24} & \textbf{18.70}& \textbf{38.15} \\
    Split Encoders &70.22  & 73.15 & 39.78 & 17.87 & 37.01\\
    QA-Span & \textbf{74.15} & \textbf{76.60}  & 41.13 &18.58& 38.04 \\
    Auto-regressive & 72.78& 74.42  & 41.04 &18.48& 37.95 \\
    \midrule
     BertSumAbs &  72.68& 76.76 &\textbf{ 41.67} & \textbf{19.46}& \textbf{38.79}  \\
     Split Encoders &72.13  & 76.43 & 40.21 & 18.38 & 37.87\\
    QA-Span & \textbf{74.93} & \textbf{78.69}  & 41.53 &19.28&38.65   \\
    Auto-regressive & 74.34 & 77.58 & 41.45 &19.18  & 38.57  \\
    
    \midrule
    BertSumExtAbs & 74.15  &79.22  & \textbf{41.87}& \textbf{19.41}&\textbf{38.94}  \\
    Split Encoders & 73.67 & 79.12 & 40.55 &    18.41&  38.45 \\
    QA-Span & \textbf{75.94} &  \textbf{80.97} & 41.75 & 19.27 &38.81  \\
    Auto-regressive &75.19  & 79.89&41.68  &19.16  & 38.74  \\
    
      \midrule
    TransformerAbs & 73.79  & 80.51 & \textbf{39.96} & \textbf{17.63} &\textbf{36.90} \\
    Split Encoders & 73.11& 79.54 & 38.83 &16.51 & 35.71 \\
    QA-Span & \textbf{75.70} & \textbf{82.82} & 39.87  & 17.50 & 36.80  \\
    Auto-regressive & 75.21& 81.64& 39.81 &17.40  &36.75  \\ 
    
    \bottomrule
  \end{tabular}
  }
  \caption{Factual correctness scores and ROUGE scores on CNN/DailyMail test set.}\label{table:cnndm_results}
\end{table}

In addition, very recent work proposed QA-based models for factuality evaluation \citep{wang2020asking,durmus2020feqa,maynez2020faithfulness}, and \citet{wang2020asking} showed that their evaluation models have higher correlation with human judgements on factuality when compared with FactCC \citep{kryscinski2019evaluating}. We thus include our re-implementation of a question generation and question answering model (QGQA) following \citet{wang2020asking} as an evaluation metric for factuality.\footnote{We were not able to obtain any of the QA evaluation model or code from \citet{wang2020asking,durmus2020feqa,maynez2020faithfulness} as the authors are still in the stage of making the code public. We used pre-trained UniLM model for question generation (QG) and BertForQuestionAnswering model for question answering (QA). The QG model is fine-tuned on NewsQA \citep{trischler2017newsqa} with entity-answer conditional task \citep{wang2020asking}, and the QA model is pre-trained on SQuAD 2.0 \citep{rajpurkar2018know}.}  This model generates a set of questions based on the system-generated summary, and then answers these questions using either the source or the summary to obtain two sets of answers. The answers are compared against each other using an answer-similarity metric (token-level F1), and the averaged similarity metric over all questions is used as the QGQA score. Answers generated from a highly faithful system summary should be similar to those generated from the source.

\subsection{Baselines}
We compare against the following abstractive summarization baselines. On CNNDM and XSum, we use  BertSumAbs, BertSumExtAbs and TransformerAbs \citep{liu2019bertabs}. In addition, we also compare with Bottom-up \citep{gehrmann2018bottom}. On Gigaword, we use the pointer-generator \citep{see2017get}, base and full GenParse models \citep{song2020joint} for comparison. For the factual correction baseline, we compare with the Two-encoder Pointer Generator\footnote{\url{https://github.com/darsh10/split_encoder_pointer_summarizer}} (Split Encoder) \citep{shah2020automatic}, which employs a similar setting to ours for masking entities w.r.t. the source, and uses dual encoders to copy and generate from both the source and the masked query for fact update. Compared to our span selection models that can fill in the mask with any number of tokens, their models aim to regenerate the mask query based on the source. In other words, their decoder regenerates the whole sequence token by token with a pointer-generator, which inherits the backbone of generative models that suffer from hallucination.

\begin{table}[t!]
\centering
\resizebox{\columnwidth}{!}{%
  \begin{tabular}{ l | c|c| c c c } 
    \toprule
     \multirow{2}{*}{Datasets} & QGQA & FactCC& \multicolumn{3}{c}{ROUGE} \\
    & & sent& 1 & 2 & L \\
    \midrule
    BertSumAbs & 12.78 & 23.60 &\textbf{37.78} & \textbf{15.84}& \textbf{30.50} \\
     Split Encoders  & \textbf{24.65} & 24.19 & 34.22 & 13.76 & 27.86 \\
    QA-Span & 23.85 & 23.90 & 36.44 &14.56& 29.38\\
    Auto-regressive & 24.14  &  \textbf{25.08} &  36.24 & 14.37 &29.22    \\
    
    \midrule
    BertSumExtAbs & 13.62  & 23.12 & \textbf{38.25}& \textbf{16.16}&\textbf{30.87}  \\
    Split Encoders &\textbf{25.17} &    24.67 & 35.66 & 13.98& 27.93\\
    QA-Span & 24.52  &23.96 &  36.86 &14.82&29.70   \\
    Auto-regressive &24.96 & \textbf{25.10} &36.67 & 14.64 & 29.53  \\
    
     \midrule
    TransformerAbs & 7.00  & 24.15 & \textbf{29.86}& \textbf{10.05} & \textbf{23.78} \\
    Split Encoders & 11.77 & 24.78  & 28.14  & 8.65& 22.70\\
    QA-Span & 12.88& 24.44 & 29.51  & 9.67 & 23.45  \\
    Auto-regressive &\textbf{ 13.89} & \textbf{25.75} &29.45 &9.59&23.40   \\

    \bottomrule
  \end{tabular}
  }
  \caption{Factual correctness scores and ROUGE scores on XSum test set.}\label{table:xsum_results}
\end{table}
\subsection{Experimental Results}

Tables~\ref{table:cnndm_results}, \ref{table:xsum_results}, and \ref{table:gigaword_result} summarize the results on the CNN/DailyMail, XSum and Gigaword datasets, respectively. Each block in the tables compares the original summarization model's output with the corrected outputs obtained by our baseline and proposed models.

On CNN/DailyMail (Table \ref{table:cnndm_results}), our correction models significantly boost factual consistency measures (QGQA and FactCC) by large margins, with only small drops on ROUGE. This shows our models have the ability to improve the correctness of system-generated summaries without sacrificing informativeness. When comparing our two proposed models, we observe that the QA-span model performs better than the auto-regressive model. 
This is expected as CNN/DailyMail's reference summaries tend to be more extractive \citep{see2017get}, and summarization models tend to make few errors per summary \citep{narayan2018don}. Thus, the iterative procedure of the QA-span model is more robust with high precision as it has more correct context from the query, with only minimum negative influence from other concurrent errors. 
This is also reflected in the high scores of QGQA and FactCC across all the models we tested. Since QGQA and FactCC are based on comparing system-generated summary w.r.t. the source text, high score means high semantic similarity between system summary to the source. 

\begin{table}[t!]
\centering
\resizebox{\columnwidth}{!}{%
  \begin{tabular}{ l | c|c| c c c } 
    \toprule
     \multirow{2}{*}{Datasets} & QGQA & FactCC& \multicolumn{3}{c}{ROUGE} \\
    & & sent& 1 & 2 & L \\
    \midrule
    GenParse (base)      &  52.63  & 46.07 & \textbf{35.23}& \textbf{17.11} & \textbf{32.88} \\
    Split Encoders & 63.60 &  48.22 & 34.32 &  17.01 & 31.98 \\
    QA-Span  &  \textbf{66.47}  &  \textbf{52.17}   & 34.38 & 16.50 & 32.07  \\
    Auto-regressive &  64.77  &  48.95  & 33.97 & 16.08 & 31.70   \\
    
    \midrule
     GenParse (full)        &  55.47  & 48.44  & \textbf{36.61} &\textbf{18.85}  & \textbf{34.32} \\
    Split Encoders & 65.88& 52.11 & 35.01 & 17.54 & 32.96\\
    QA-Span  & \textbf{67.12}  & \textbf{54.59} &35.66  & 18.01  & 33.35   \\
    Auto-regressive & 66.48  & 52.18 & 35.04  & 17.27 & 32.75  \\
    
     \midrule
    Pointer Generator         & 45.98  & 43.62 & \textbf{34.19}  &\textbf{16.29}  & \textbf{31.81} \\
    Split Encoders & 59.46 & 48.32 & 33.11 & 15.63& 30.67 \\
    QA-Span  &  58.25  &    45.62  & 33.30 & 15.70 &  30.95  \\
    Auto-regressive  & \textbf{60.66}  & \textbf{49.82} &32.86  & 15.22 &  30.51\\
    \bottomrule
  \end{tabular}
  }
  \caption{Factual correctness scores and ROUGE scores on Gigaword test set.}\label{table:gigaword_result}
\end{table}

On XSum (Table~\ref{table:xsum_results}) and Gigaword (Table~\ref{table:gigaword_result}), both of our correction models boost factual consistency measures by large margins with a slight drop in ROUGE (-0.5 to -1.5) on average. This is still encouraging, as abstractive summarization models that use complex factual controlling components for generation often have drops of 5-10 ROUGE points \citep{zhu2020boosting}. 


We also notice that the QGQA and FactCC scores of all summarization models are lower than that on CNN/DailyMail. The scores are especially low on XSum. This is likely due to the data construction protocol of XSum, where the first sentence of a source document is used as the summary and the remainder of the article is used as the source. As a result, many entities that appear in the reference summary never appear in the source, which may cause abstractive summarization models to hallucinate severely with many factual errors \citep{maynez2020faithfulness}.  As the system summaries often contain many errors, our QA-span model that relies on answering a single-mask query often has the wrong context to condition on at each step, which negatively affects the performance of this model. In contrast, the strategy of masking all the entities would provide the auto-regressive model a better query for entity replacement. We can observe in Table \ref{table:xsum_results} that the auto-regressive model performs better than the QA-span model on XSum.

\subsection{Human Evaluation}

\begin{table}[t!]
    \centering
    \resizebox{\columnwidth}{!}{
    \begin{tabular}{c|cc|c}
    \toprule
         BertAbs & Better & Worse & Same \\\midrule
         QA-Span vs. original & \textbf{28.6\%} &  18.7\% & 52.7\% \\
         Auto-regressive vs. original & \textbf{31.3\%}  & 16.7\% & 52\%\\
        QA-Span vs. Auto-regressive & 26\%  & 27.3\% & 46.7\%\\
         \midrule
         
         TransformerAbs & Better &   Worse & Same\\\midrule
        QA-Span vs. original & \textbf{38\%}  & 11.3\% & 40.7\% \\
         Auto-regressive vs. original & \textbf{36\%}  & 19.3\% & 44.7\%\\
         QA-Span vs. Auto-regressive & 32.7\%  & 28\% & 39.3\%\\
         \midrule
         
         Bottom-up & Better & Worse  & Same \\\midrule
         QA-Span vs. original & \textbf{34\%} & 12\%  & 54\% \\
         Auto-regressive vs. original & \textbf{31.4\%}  & 13.3\% & 55.3\% \\
         QA-Span vs. Auto-regressive & 41.3\%  & 32\% & 26.7\%\\
         \bottomrule
       
    \end{tabular}
    }
    \caption{Human evaluation results on pairwise
comparison of factual correctness on 450 ($9 \times 50$) randomly sampled articles.}
    \label{tab:human_eval}
\end{table}

To provide qualitative analysis of the proposed
models, we conduct human evaluation on pairwise comparison of CNN/DailyMail summaries enhanced by different correction strategies.
We select three state-of-the-art abstractive summarization models as the backbones, and collect three sets of pairwise summaries for each setting: $(i)$ Original vs. QA-Span corrected; $(ii)$ Original vs. Auto-regressive corrected; $(iii)$ QA-Span corrected vs. Auto-regressive corrected. Nine sets of 50 randomly selected samples (total 450 samples) are labeled by AMT tuckers. For each pair (in anonymized order), three annotators from Amazon Mechanical Turk (AMT) are asked to judge which is more factually correct based on the source document.  
As shown in Table \ref{tab:human_eval}, summaries from our two models are
chosen more frequently as the
factually correct one compared to the original. Between the two correction models, the preferences are comparable.

In addition, we also test our fact correction models on  the FactCC test set provided by \citet{kryscinski2019evaluating} and manually checked the outputs. Table \ref{tab:FactCC_result} shows the results of the original summaries and the summaries corrected by our models in terms of automatic fact evaluation and our manual evaluation.  Among 508 system-generated summary sentences, 62 were incorrect. The QA-span model was able to correct 18 out of 62 right, and the auto-regressive model was able to correct 16 out of 62. Among the 446 sentences that are labeled as correct by the annotators in \citet{kryscinski2019evaluating}, our two models made 3 and 4 wrong changes in the entities, respectively,\footnote{This excludes the cases where the model would change a person's full name by last name or break the fluency due to SpaCy NER errors.} while keeping most of the entities unchanged or changed with equivalent entities. 

\begin{table}[t!]
    \centering
    \resizebox{\columnwidth}{!}{
    \begin{tabular}{c|c|c|c}
    \toprule
         FactCC Dataset& FactCC Score & QAQG & Human Eval\\\midrule
         Before Corr.& 84.89 & 88.65 & 87.79 \\
         QA-span & \textbf{86.08} & \textbf{91.07} & \textbf{90.74} \\
         Auto-regressive & 85.96 & 90.51 & 90.35\\
         \bottomrule
    \end{tabular}
    }
    \caption{Test results on the human annotated dataset provided by FactCC \citep{kryscinski2019evaluating}.  We show the performance comparisons of the original summaries and the summaries corrected by SpanFact.}
    \label{tab:FactCC_result}
\end{table}

\section{Conclusion}
We present SpanFact, a suite of two factual correction models that use span selection mechanisms to replace one or multiple entity masks at a time. SpanFact can be used for fact correction on any abstractive summaries. Empirical results show that our models improve the factuality of summaries generated by state-of-the-art abstractive summarization systems without a huge drop on  ROUGE scores.  For future work, we plan to apply our method for other type of spans, such as noun phrases, verbs, and clauses. 

\section*{Acknowledgments}

This  research  was  supported  in  part  by  Microsoft
Dynamics 365 AI Research and  the Canada CIFAR AI Chair program.
We would like to thank the reviewers for their valuable comments and special thanks to Yuwei Fang and other members of the Microsoft
Dynamics 365 AI Research team for the feedback and suggestions. 

\bibliography{emnlp2020}
\bibliographystyle{acl_natbib}

\clearpage
\appendix


\begin{table*}[ht!]
\resizebox{\textwidth}{!}{
\begin{tabular}{  m{5em} |  m{14cm}  } 
\hline
CNNDM Source& Jerusalem (CNN)The flame of remembrance burns in Jerusalem, and a song of memory haunts Valerie Braham as it never has before. This year, \textcolor{blue}{Israel}'s Memorial Day commemoration is for bereaved family members such as Braham. ``Now I truly understand everyone who has lost a loved one," Braham said. Her husband, \textcolor{blue}{Philippe Braham}, was one of 17 people killed in January's terror attacks in Paris. He was in a kosher supermarket when a gunman stormed in, killing four people, all of them Jewish. \\ 
\hline
 System Summary & \textcolor{red}{france}'s memorial day commemoration is for bereaved family members as braham. \textcolor{red}{valerie braham} was one of 17 people killed in january's terror attacks in paris.\\
\hline
Corrected by SpanFact & \textcolor{orange}{israel}'s memorial day commemoration is for bereaved family members as braham. \textcolor{orange}{philippe braham} was one of 17 people killed in january's terror attacks in paris.\\
\hline \hline
CNNDM Source& (CNN)If I had to describe the U.S.-Iranian relationship in one word it would be ``overmatched." ... \textcolor{blue}{America} is alienating some of our closest allies because of the Iran deal, and Iran is picking up new ones and bolstering relations with old ones who are growing more dependent because they see Iran\'s power rising...  \\ 
\hline
 System Summary & \textcolor{red}{iran} is alienating some of our closest allies because of the iran deal, and iran is picking up new ones.\\
\hline
Corrected by SpanFact & \textcolor{orange}{america} is alienating some of our closest allies because of the iran deal, and iran is picking up new ones.\\
\hline \hline
CNNDM Source& (CNN)A North Pacific gray whale has earned a spot in the record books after completing the longest migration of a mammal ever recorded. The whale, named Varvara, swam nearly 14,000 miles (22,500 kilometers), according to a release from \textcolor{blue}{Oregon State University}, whose scientists helped conduct the whale-tracking study. Varvara, which is Russian for ``Barbara," left her primary feeding ground off \textcolor{blue}{Russia\'s Sakhalin Island} to cross the  Pacific Ocean and down the West Coast of the United States to Baja, Mexico... \\ 
\hline
 System Summary & a north pacific gray whale swam nearly 14,000 miles from \textcolor{red}{oregon state university}.\\
\hline
Corrected by SpanFact & a north pacific gray whale swam nearly 14,000 miles from \textcolor{orange}{russia\'s sakhalin island}.\\
\hline \hline
CNNDM Source& Sanaa, Yemen  (CNN)Saudi airstrikes over Yemen have resumed once again, \textcolor{blue}{two days after} Saudi Arabia announced the end of its air campaign. The airstrikes Thursday targeted rebel Houthi militant positions in three parts of Sanaa, two Yemeni Defense Ministry officials said. The attacks lasted four hours. ... The Saudi-led coalition said a new initiative was underway, Operation Renewal of Hope, focused on the political process. But \textcolor{blue}{less than 24 hours later}, after rebel forces attacked a Yemeni military brigade, the airstrikes resumed, security sources in Taiz said.\\ 
\hline
 System Summary & the attacks lasted four hours, \textcolor{red}{two days after} rebel forces attacked yemeni military troops..\\
\hline
Corrected by SpanFact & the attacks lasted four hours, \textcolor{orange}{less than 24 hours after} rebel forces attacked yemeni military troops.\\
\hline \hline
CNNDM Source& Boston (CNN)When the bomb went off, \textcolor{blue}{Steve Woolfenden} thought he was still standing. That was because, as he lay on the ground, he was still holding the handles of his son's stroller. He pulled back the stroller's cover and saw that his son, \textcolor{blue}{Leo}, 3, was conscious but bleeding from the left side of his head. Woolfenden checked Leo for other injuries and thought, ``Let's get out of here." ...\\ 
\hline
 System Summary & \textcolor{red}{steve woolfenden}, 3, was conscious but bleeding from the left side of his head.\\
\hline
Corrected by SpanFact & \textcolor{orange}{leo}, 3, was conscious but bleeding from the left side of his head.\\
\hline
\end{tabular}
}

\caption{Examples of factual error correction on FactCC dataset (a human annotated  subset from CNNDM obtained by \citet{kryscinski2019evaluating}).  Factual
errors by abstractive summarization system are marked in red. Corrections made by the proposed SpanFact models are marked in orange.}
\end{table*}

\end{document}